\documentclass[runningheads]{llncs}
\usepackage{graphicx}

\usepackage{float}
\usepackage{moresize}
\usepackage{multirow}
\usepackage{booktabs}
\usepackage{amsmath}
\usepackage{adjustbox}
\usepackage{amsfonts}
\usepackage{subfig}

\usepackage{amssymb}

\usepackage{mathrsfs}

\usepackage{array}
\newcolumntype{R}[1]{>{\raggedleft\let\newline\\\arraybackslash\hspace{0pt}}m{#1}}
\newcolumntype{L}[1]{>{\raggedright\let\newline\\\arraybackslash\hspace{0pt}}p{#1}}
\newcolumntype{C}[1]{>{\centering\let\newline\\\arraybackslash\hspace{0pt}}m{#1}}
\usepackage{textcomp}

\usepackage{threeparttable}
\newtheorem{mydefinition}{Definition}

\begin{document}

%
%

\institute{}

\authorrunning{First Author et al.}
%
\title{Multivariate Business Process Representation Learning utilizing Gramian Angular Fields and Convolutional Neural Networks
}
\titlerunning{MPPN}
%
\author{Peter Pfeiffer \and
Johannes Lahann \and
Peter Fettke}
%


\institute{German Research Center for Artificial Intelligence (DFKI) and Saarland University, Saarbrücken, Germany\\
\email{\{peter.pfeiffer, johannes.lahann, peter.fettke\}@dfki.de}}
\authorrunning{Pfeiffer et. al.}
\maketitle              
\begin{abstract}
Learning meaningful representations of data is an important aspect of machine learning and has recently been successfully applied to many domains like language understanding or computer vision. Instead of training a model for one specific task, representation learning is about training a model to capture all useful information in the underlying data and make it accessible for a predictor. For predictive process analytics, it is essential to have all explanatory characteristics of a process instance available when making predictions about the future, as well as for clustering and anomaly detection. Due to the large variety of perspectives and types within business process data, generating a good representation is a challenging task.
In this paper, we propose a novel approach for representation learning of business process instances which can process and combine most perspectives in an event log. In conjunction with a self-supervised pre-training method, we show the capabilities of the approach through a visualization of the representation space and case retrieval. Furthermore, the pre-trained model is fine-tuned to multiple process prediction tasks and demonstrates its effectiveness in comparison with existing approaches.

\end{abstract}

\section{Introduction}

Current machine-learning-based methods for predictive problems on business process data, e.g., neural-network-based methods like LSTMs or CNNs, achieve high accuracies in many tasks such as next activity prediction or remaining time prediction on many publicly available datasets \cite{Neu_2021}.
In recent time, a large variety of new architectures for next step and outcome prediction have been proposed and evaluated \cite{Pasquadibisceglie_2020_2,Pasquadibisceglie_2021,Farbod_2020}. 
These machine-learning-based methods are mostly task-specific and not generic, i.e., they are designed and tested on predictive process analytics tasks like next step prediction, outcome prediction, anomaly detection, or clustering. Moreover, most of the proposed approaches process only one or a limited set of attributes including activity, resource \cite{Evermann_2017} or timestamp \cite{Tax_2017}. On one hand, predicting the next activity in an ongoing case using the control-flow information is very similar to predicting the next word in a sentence. On the other hand, data in an event log recorded from business process executions is very rich in information.
Usually, there are various attributes with different types, scales, and granularity, which makes generating good representations containing all characteristics a difficult tasks. Characteristics of a process instance that are embedded in the data attributes, can improve the performance of predictive models \cite{Marquez_2018}. For example, the next activity in a business process can depend on multiple attributes in previous events and complex dependencies between these attributes \cite{Brunk_2020}.


Inspired by recent research in the language modeling domain \cite{Devlin_2018_BERT}, we propose and evaluate a novel and generic network architecture and training method -- the Multi-Perspective Process Network (MPPN) -- which learns a meaningful, multi-variate representation of a case in a general-purpose feature vector that can be used for a variety of tasks. The research contributions of this paper is threefold:
\begin{enumerate}
    \item We introduce a novel neural-network-based architecture to process a flexible number of process perspectives of a process instance that examines all of its characteristics using gramian angular fields.
    \item For this architecture, we propose an self-supervised pre-training method to generate a feature-vector representation of a process instance that can be fine-tuned to various tasks, thus making a contribution to representation learning for predictive process analytics.
    \item We show the effectiveness of this approach by analyzing the representation space and performing an unsupervised case-retrieval task. Furthermore, we fine-tune and compare the model on a variety of predictive process analytic tasks such as next step and outcome prediction against existing approaches.
\end{enumerate}
The structure of the remaining chapters unfolds as follows: Section 2 introduces the reader to preliminary concepts.
Section 3 discusses related work on the use of machine learning in predictive process analytics and representation learning for business process data. 
Section 4 and 5 present the proposed approach and the evaluation on a variety of predictive process analytic tasks. Section 6 closes the paper with a summary of the main contributions and findings as well as an outline of future work.

\section{Foundations}
\subsection{Business Process Event Log and Perspectives}
Event logs contain records from process-aware information systems in a structured format. These recordings contain information about what activities have been conducted by whom at what time as well as additional contextual data. The following definitions will be used in later sections of the paper.

\begin{mydefinition}{Event Log}\\
An event log is a tupel $L=(E,<,V_1,\dots,V_n,a_1,\dots,a_n)$, where
\begin{itemize}
    \item E is the set of events
    \item $<$ is a total order on E
    \item $V_1,\dots,V_n$ are the sets of the attribute values
    \item attributes $a_i:E \rightarrow V_i$, maps an event to an attribute value.
\end{itemize}
\end{mydefinition}
In the following, we expect an event log to have at least the attributes \textit{case-id}, \textit{activity}, \textit{resource}, and \textit{timestamp}.

\begin{mydefinition}{Case}\\
Let $a_j$ be the attribute \textit{case-id} and $v \in V_j$. C is the set of events of one case, iff
\begin{enumerate}
    \item $C \subseteq E$
    \item For each $e \in C: a_j(e)=v$
    \item For each $e \in E \setminus C : a_j(e) \neq v$
\end{enumerate}
\end{mydefinition}
Let $C= \{e_1,\dots,e_n\} $ be the events of a case $c$. These events follow the order $<$. We use the notation: $c=<e_1,\dots,e_n>$, where $e_i<e_j$ if $i<j$.   

\begin{mydefinition}{Business Process Perspective}\\
Given $L=(E,<,V_1,\dots,V_n,a_1,\dots,a_n)$ and a sequence of events $ \langle e_1,\dots,e_n \rangle$, we define a perspective on each attribute $a_i$ as $\Pi_{a_i}:=\langle a_i(e_1),\dots,a_i(e_n) \rangle$
\end{mydefinition}
In the next sections, we frequently use the control-flow perspective $\Pi_{control-flow}$, the resource perspective $\Pi_{resource}$ and the temporal perspective $\Pi_{time}$.
\newline

\noindent Last, we distinguish between event attributes and case attributes. A case attribute returns the same value for all events in a case, i.e. it fulfils equation 2 in definition 2. Otherwise, it is an event attribute.

\subsection{Business Process Data and Representation Learning}
Representation learning is the task of learning "representations of data that make it easier to extract useful information when building classifiers or other predictors" \cite{Bengio_2013}. For example, embeddings are utilized in natural language processing to learn a vectorized representation of words. In recent times, attention-based networks \cite{Vaswani_2017} have shown superior performance in many language tasks such as machine translation, mainly due to their ability to generate meaningful representations. Usually, these types of networks are pre-trained in an unsupervised fashion on extensive datasets.
For business process data, embeddings are commonly used in predictive process analytics to represent the control-flow \cite{Camargo_2019}, or certain perspectives \cite{Evermann_2017} in a vector space. Thus, it should allow the model to exploit the vector representation more effectively than a one-hot or integer encoding. When learning good representations of cases, one tries to represent all relevant characteristics within the representation. For predictive tasks, this includes the underlying distribution of the exploratory factors, i.e., attributes that influence the prediction \cite{Brunk_2020}. 
Usually, there are several attributes with different types of data -- categorical, numerical, and temporal ones. Within each type, the dimensions, scales, and variabilities can be different. For instance, there can be a numerical attribute \textit{cost} ranging from [0, 1,000,000] and another one, e.g., \textit{discount} that is in range [0, 30]. Temporal attributes also have different scales (daily, weekly, monthly, etc.) and high variability. Categorical attributes often vary in their dimensionality. While the \textit{activity} has few different values, the \textit{resource}, e.g., persons involved in a process, often has much more distinct values. Some perspectives are very spare while others are rich. Furthermore, different from event-related attributes, there are also case-related attributes. Some attributes change their values only in certain events of a case, while others have a different value for each event. This, in turn, means that perspectives have different levels of granularity. When learning representations of cases, the multivariate, multi-scalar, and multi-granular nature of the data must be considered and depicted.

\section{Related Work} \label{section:related_work}
In \cite{Weerdt_2018}, the authors introduced methods to learn representations of activities, cases, and process models. They trained a model on the next step prediction task to learn representations similar to obtaining embeddings for words, sentences, and documents. Although they explained that other attributes besides activity are important, they only consider the control-flow perspective. Furthermore, they did not include an extensive evaluation to elaborate on the effectiveness of the proposed representations.
Apart from that, representation learning is not explicitly tackled in existing predictive analytic approaches. However, these approaches learn a representation alongside a specific prediction task. 
\cite{Evermann_2017} was the first to introduce neural networks to the field of process prediction. They applied recurrent neural networks to next activity and remaining time prediction. They trained separate neural networks considering the control-flow, resource, and time perspectives. 
\cite{Tax_2017} examined the next step and remaining time prediction task. They used an LSTM-based approach that optimized both tasks simultaneously and elaborated on the effect of separated or shared LSTM network layers.
\cite{Camargo_2019} elaborated three different LSTM architectures for predicting the next activity, resource, and timestamp. In their first architecture, they used specialized layers for each attribute to predict. The second version combines the categorical attributes in a shares layer, while the third version shares categorical attributes and the timestamp. 
In \cite{Pasquadibisceglie_2021}, the authors proposed an LSTM-based method that can combine multiple attributes. In their model, the authors use embeddings for each categorical attribute while non-categorical attributes are concatenated with the categorical ones' embedded representation. 
Other approaches for next step prediction used different techniques such as decision trees (\textit{DT}) \cite{Maggi_2014}, autoencoder (\textit{AE}) with n-grams \cite{Mehdiyev_2020}, attention networks \cite{Moon_2021}, CNNs \cite{Pasquadibisceglie_2020} or generative adversarial networks (\textit{GAN}) \cite{Farbod_2020}. Similarly, CNNs \cite{Pasquadibisceglie_2020_2}, LSTMs \cite{Navarin_2017} or autoencoder \cite{Mehdiyev_2020_2} are used for outcome prediction. 
In order to detect anomalies in business process data, LSTMs \cite{Nolle_BINet} or Bayesian neural networks \cite{Pauwels_2019} were applied. 
Table \ref{tab:encoding_techniques} gives an overview of existing predictive approaches and categorizes them by prediction task, examined perspective, encoding technique per perspective, as well as used machine learning method. Also, it clarifies what information is available to which model in what form. While all predictive approaches use at least information from two perspectives, only a few approaches are able to encode and process all types of attributes. We differentiated between the most commonly used perspectives and other attribute types to delimit generic and non-generic approaches. Only \cite{Pasquadibisceglie_2021} used a generic encoding approach that can process and represent all types of attributes. However, the approach is tailored towards next step prediction and does not focus on the learned representation within the model.
Thus, we propose a generic multi-attribute representation learning approach that is not tailored to a specific prediction task.  
\newcommand{\ignore}[1]{}

\begin{table}[bt]
\centering
\caption{Overview of encoding techniques used in literature. \textit{duration}: timestamps to duration after a certain timestamp, \textit{IMG}: one or multiple perspectives $\Pi$ to a single or multiple matrices $M^{h \times w \times c}$ of size $height \times width \times \#channels$.}
\resizebox{\textwidth}{!}{
\begin{tabular}{p{2cm} l p{2.5cm} p{2cm} p{2cm} p{2cm} p{2cm} p{2cm} p{2cm} }
\toprule
\multirow{3}{*}{Task}                 & \multirow{3}{*}{Approach} & \multirow{3}{*}{ML Method} 
& \multicolumn{6}{c}{Encoding Techniques for event attributes} \\ \cmidrule(lr){4-9} 
        & & & \multicolumn{3}{c}{Most common perspectives} & \multicolumn{3}{c}{Other event log attributes} \\ \cmidrule(lr){4-6} \cmidrule(lr){7-9}

        & & & $\Pi_{control-flow}$& $\Pi_{resource}$& $\Pi_{timestamp}$   & Categorical    & Numerical         & Temporal \\ \midrule

\multirow{9}{*}{\parbox{2cm}{Next\\ Step\\ Prediction}}   & \ignore{Evermann} \cite{Evermann_2017}     & LSTM
            & \multicolumn{2}{c}{embedding}    & -           & -              & -                 & -    \\ \cline{2-9}

                                        & \ignore{Tax} \cite{Tax_2017}               & LSTM               
            & one-hot           & -            & custom                 & -              & -
                & -      \\ \cline{2-9} 
                                      
                                        & \ignore{Mehdiyev} \cite{Mehdiyev_2020}     & AE + FF
            & N-gram           & one-hot             & -                 & one-hot             & as-is                 & -      \\ \cline{2-9} 
                                      
                                        & \ignore{Maggi} \cite{Maggi_2014}           & DT                           
            & integer                &               &                  & integer              &                   &       \\ \cline{2-9}
                                        
                                        
                                        & \ignore{PREMIER} \cite{Pasquadibisceglie_2020} & CNN
            & \multicolumn{3}{c}{IMG}         & -             & -                 & -   \\ \cline{2-9} 
            
                                        & \ignore{POP-ON} \cite{Moon_2021} & Attention
            & one-hot           & one-hot             & -         & one-hot         & as-is             & -   \\ \cline{2-9}
                                        
                                        & \ignore{MiDA} \cite{Pasquadibisceglie_2021} & LSTM
            & embedding         & embedding  & duration             & integer               & as-is     & duration  \\ \cline{2-9}
            
                                        & \ignore{Camargo} \cite{Camargo_2019}  & LSTM
            & embedding         & embedding       & duration         & -                     & -         & -  \\ \cline{2-9}
            
                                        & \ignore{Farbod} \cite{Farbod_2020}  & GAN
            & one-hot         & -       & duration         & -                     & -         & -  \\ \midrule
            
                                        
\multirow{2}{*}{\parbox{2cm}{Outcome\\ Prediction}}     & \ignore{ORANGE} \cite{Pasquadibisceglie_2020_2} & CNN
            & \multicolumn{3}{c}{IMG} & -             & -                 & -       \\ \cline{2-9} 
            
            
                                        & \ignore{Mehdiyev} \cite{Mehdiyev_2020_2}   & FF              
            & custom           & custom           & custom                & custom      & custom               & custom   \\ \cline{2-9}
                                        
                                        & \cite{Navarin_2017}           & LSTM                                      
            & one-hot           & one-hot       & custom           & one-hot        & -                 & -  \\ \midrule
                                        

                                        
\multirow{2}{*}{\parbox{2cm}{Anomaly\\ Detection}}     & \ignore{BINet} \cite{Nolle_BINet}        & LSTM 
            & integer       & integer             & -               & integer       & -              & -     \\ \cline{2-9}
            
                                                    & \cite{Pauwels_2019} & Bayesian NN           
            & probability   & probability   & probability           & probability   & probability    & probability     \\ \midrule
                                        
All                                                 & MPPN                          & CNN                                              
            & IMG           & IMG           & IMG         & IMG     & IMG           & IMG   \\ \bottomrule
\end{tabular}}
\label{tab:encoding_techniques}
\end{table}

\section{Multi-Perspective Process Network (MPPN)}
\begin{figure}[bt]
    \centering
    \includegraphics[width=0.70\textwidth]{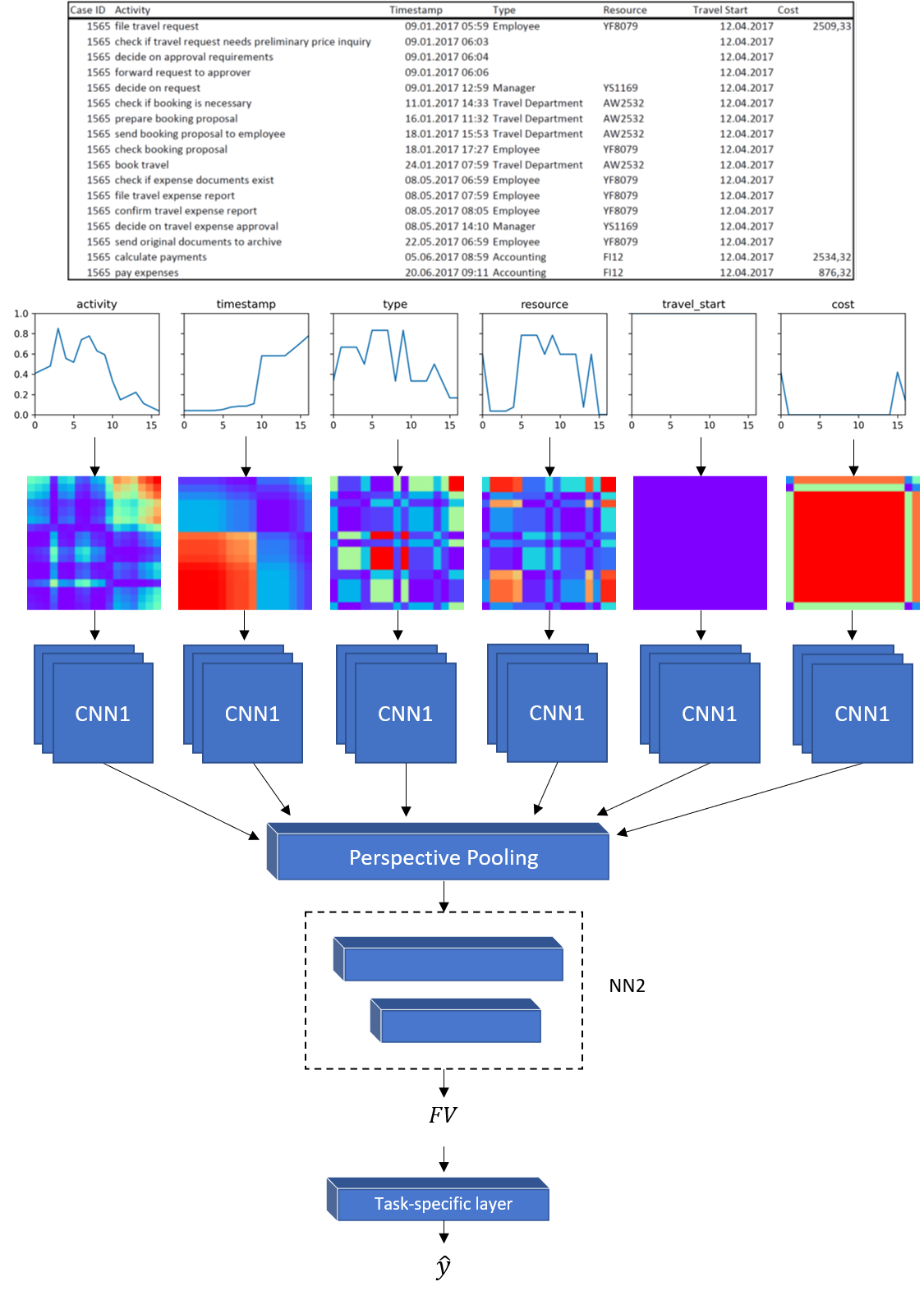}
    \caption{Architecture and processing pipeline of MPPN using a case from the MobIS event log \cite{Houy_2019}.}
    \label{fig:MPPN_scheme}
\end{figure}

The MPPN approach for representation learning is mainly built on two concepts -- graphical event log encoding and neural-network-based processing. The first part is to encode the perspectives of interest $\hat{\Pi}$ in the event log $L$ uniformly as 2D images by transforming them to distinct gramian angular fields (GAF) -- no matter if the perspective contains categorical, numerical or temporal information. The second part is a convolutional neural network architecture and training method that learns representations of cases using the GAF-encoded perspectives.
Figure \ref{fig:MPPN_scheme} shows the architecture and processing pipeline of the MPPN approach. In this example, the six perspectives of interest $\hat{\Pi} = \{\Pi_{control-flow},\\ \Pi_{timestamp}, \Pi_{type}, \Pi_{resource}, \Pi_{travel\_start}\, \Pi_{cost}$ of the case with \textit{case-id} 1565 are first encoded as six individual GAFs, which can then be processed by the MPPN. 
CNN1 extracts the features of each perspective $\hat{\Pi}$ which are combined in the perspective pooling layer. The combined features capture all characteristics of interest of a particular case.
The forwards pass in MPPN of a single case $c$ is as follows: For each GAF-encoded perspective, \textit{CNN1} extracts a feature vector that is pooled before being passed to \textit{NN2}. \textit{NN2} then takes the pooled features from all perspectives, processes them, and produces a single feature vector $FV$. This two-stage architecture allows \textit{CNN1} to focus on the features within each perspective while \textit{NN2} captures and models the dependencies between perspectives. By transforming all perspectives uniformly to GAFs, all attributes lie within the same range, no matter what scale or variability they had before. At the same time, \textit{NN2} can learn what features from what perspectives are important. Unlike RNN-based models, MPPN consumes the whole case at once instead of being fed with cases event by event.\\
Inspired by multi-view learning, e.g., Multi-View Convolutional Neural Networks (MVCNN) for 3D object detection using 2D renderings \cite{Learned-Miller_2015}, the Multi-Perspective Process Network creates a feature vector $FV$ for a case using the GAF-encoded perspectives. Analogous to using multiple renders from different views to represent 3D structures, we use different perspectives of a process to represent a case. Another important aspect of MPPN is its ability to be used for several tasks instead of being task-specific. This is achieved by an self-supervised pre-training phase, as also done in\cite{Mehdiyev_2020}, that learns a representation in the form of a feature vector $FV$. Afterwards, a task-specific layer can be added to the pre-trained model allowing to fine-tune the model on different tasks. The learned representation $FV$ thus serves as the basis for any downstream task.

\subsection{Graphical Representation of Event Log Data}
In order to encode all types of attributes into a single representation in a generic way, we decided to choose a graphical encoding instead of the methods used in related work shown in table \ref{tab:encoding_techniques}. We see a strong similarity in the characteristics of time-series and a single perspective of a process case. Furthermore, all types of attributes in a case can easily be transformed to time-series. For this reason, we treat the perspectives $\hat{\Pi}$ as multivariate time series.
A naive way to get a both machine-readable and visualizable representations of perspectives $\Pi$ is to represent and plot them as time series. 
Thus, each value of a perspective is encoded as a real number and visualized in a 2D representation. The y-coordinate corresponds to the value $v$, and the x-coordinate to $t$. In figure \ref{fig:MPPN_scheme}, one can see the 6 perspectives $\hat{\Pi}$ of case 1565 encoded and plotted as 6 distinct time series. 
Although this representation is a nice visualization for humans, presenting the perspectives $\Pi$ as a time series plot is a very naive way. Such a plot is very sparse, i.e., most of the plot is empty with just a fine line drawn, containing only little information for convolutional neural networks.

Gramian angular fields (GAF), originally proposed for time-series classification, transform sequential data to 2D images, which contain more information for machine learning methods as time-series plots \cite{Wang_2014}. 
For a sequence $ \langle v_{1}, ..., v_n \rangle$, a gramian angular field is a matrix of size $n \times n$ where each entry is the cosine of the sum of two polar coordinates in time -- the polar coordinate of $v_{i}$ plus $v_{j}$. This projection is bijective and preserves temporal relations. To transform event log data, i.e., all perspectives $\hat{\Pi}$ of categorical, numerical, and temporal event log attributes and case attributes into gramian angular fields, they must be treated and transformed to distinct sequences of numerical values. In order to get numerical sequences from each type, the following transformations and encodings are performed. Other types of attributes can also be used (e.g., textual data) if they are encoded as numerical sequences.
\begin{enumerate}
    \item For categorical attributes $a_i$, we applied an integer encoding $integer: \mathcal{V}_{i} \rightarrow int$ where $int \in [0, 1, 2, .., |\mathcal{V}_{i}|-1]$.
    \item Timestamps are transformed into the duration in seconds from the earliest timestamp.
    \item Numerical attributes are used unchanged.
\end{enumerate}
Case attributes are first duplicated to the case length before being encoded in the same way as event attributes.
Once encoded as numerical sequences, each perspective can easily be encoded as a gramian angular field after scaling them to a [-1, 1] range. To ensure equal size images where characteristics are equally represented, the sequences are adjusted to equal length, either by padding or truncating. This results in distinct GAF-representations for each perspective $\hat{\Pi}$ of a case as shown in figure \ref{fig:MPPN_scheme}.\\
By using graphical encodings we can transform attributes of different types to images and use state-of-the-art image processing neural networks. This way we avoid building networks that process cases with customized architectures for specific attribute types as well as the complexity of training embeddings.

\subsection{Architecture}
The MPPN architecture consists of three parts as shown in figure \ref{fig:MPPN_scheme}: \textit{CNN1} for feature extraction, \textit{NN2} for modeling dependencies and relations between perspectives, and one or multiple task-specific layers called \textit{HEAD}. Between \textit{CNN1} and \textit{NN2}, a pooling layer combines the features produced by \textit{CNN1} for each GAF-encoded perspective to a single vector. The weights in \textit{CNN1} are shared between all perspectives, i.e. the same \textit{CNN1} is applied on all perspectives.
For \textit{CNN1}, we use Alexnet \cite{Alex_2012}.
However, as gramian angular fields are different from natural images, pre-training \textit{CNN1} on GAFs significantly reduces the later training time. \textit{NN2} is a fully-connected neural network. Together, \textit{CNN1} and \textit{NN2} form the model used for representation learning that produces $FV$. $FV$ can either be used directly or by any other task-specific layer; e.g., a fully-connected \textit{HEAD} with \textit{softmax} for next step prediction or a \textit{HEAD} for remaining time prediction.

\subsection{Training Method}
One integral part of MPPN is its ability to learn representations of all perspectives $\hat{\Pi}$ of cases in an event log. In order to obtain good feature vectors $FV$, one must ensure that all relevant characteristics are fully captured in the model. We distinguish three stages of training that should be performed successively. 

\subsubsection{Pre-Training CNN1 on GAFs}
As GAFs are very different from natural images, pre-trained CNNs like Alexnet need to be fine-tuned. While lower-level features like edges and corners are present in GAFs too, higher-level features differ. In order to make the MPPN sensitive to GAF-specific feature, we fine-tuned the \textit{CNN1} once by classifying cases according to their variant. This task has been chosen as the process variant is always directly derivable from the sequence of events and the model can focus on learning features from the single GAF-encoded perspective. All relevant information for this task is entailed in the GAF image which is what we want the model to focus on. However, many other tasks are also possible.
In detail, we build a MPPN with a pre-trained Alexnet consuming only $\Pi_{control-flow}$ and predicting the variant on the MobIS dataset. 
For each case $c$ the whole sequence of $activity$ was used as input and the variant used as the target.  Afterwards, the weights of \textit{CNN1} are saved on disk and can be used on any dataset, any perspective, and any task for MPPN in the future.

\subsubsection{Representation Learning} \label{approach:representation_learning}
To obtain meaningful feature vectors $FV$ of business process cases, MPPN must be trained to hold all characteristics of a case in it. One can train MPPN on next step prediction tasks, e.g., to predict the next activity in an ongoing case given $\hat{\Pi}$. This works fine, but the model will learn the relation in the data, which are important for the next activity. This leads to a feature vector $FV$ that by design holds features that are important to predict the next activity. Attributes that do not have relevance for the next activity will be less present in $FV$.\\
To obtain more generic feature vectors of cases, a self-supervised multi-task next event prediction training method is applied that trains the network to predict $a_i(e_{t+1})$ for each attribute in $\hat{\Pi}$. For this task, the MPPN architecture is extended by small networks $HEAD_{a_i}$ -- one for each attribute $a_i$ to predict. Each $HEAD$ is a task-specific layer that consumes $FV$ and predicts $a_i(e_{t+1})$. During representation learning, the task's criterion is to minimize the sum of all losses of all predictions, measured as mean absolute error (for numerical and temporal attributes) and cross-entropy (for categorical attributes). During training, all $HEADs$ are trained in parallel and in conjunction with the rest of MPPN.
Thereby, the MPPN and especially \textit{NN2} learns to focus on important features in all perspectives $\hat{\Pi}$ and produces a $FV$ that holds information relevant for the attributes in the next event. Using this method, a representation can be learned without the need for manual labeled data. 
However, depending on the final task to be solved, other training methods are also possible. As long as all relevant characteristics of the case are enclosed in $FV$, any training method is appropriate. We chose the multi-task next event prediction task as it allows the model to incorporate all attributes for each prediction. While making predictions for each attribute the model is forced to not drop relevant characteristics of a case.
Afterwards, the weights of MPPN (without the heads) are saved on disk. The $FV$ produced in this state can directly be used for tasks where additional labels are hard to obtain or unavailable, such as clustering, retrieval or anomaly detection using the same dataset.

\subsubsection{Fine-tuning on Specific Tasks} \label{approach:fine_tuning}
After being trained to learn good representations, MPPN can also be fine-tuned on other tasks using the same event log and given appropriate labels. Therefore, one or multiple $HEADs$ are added that consume $FV$. With each \textit{HEAD}, the model and especially the \textit{HEAD} can be trained on a large variety of tasks, e.g., outcome prediction, next step prediction or (supervised) anomaly detection. Thereby, the model makes use of the representation in $FV$ to solve a certain problem.

\subsection{Implementation Details} \label{sec:implementation_details}
We implemented MPPN with the following hyperparameter choices: We padded or truncated all cases $c$ to length 64 which results in GAF images of size $64\times64$ pixel. \textit{CNN1} consists of four CNN layers with max-pooling and dropout. \textit{NN2} is a two-layer fully-connected network with dropout. We pooled the perspectives behind \textit{CNN1} by concatenation. The \textit{HEADs} consist of shallow fully-connected networks with a softmax or regression layer. More details can be found in the implementation.

\section{Evaluation}
This section elaborates on two experiments. The first experiment visualizes the learned representations during the self-supervised pre-training phase and demonstrates a contextual retrieval task. In the second experiment, we compare the MPPN model to existing approaches on next event and outcome prediction tasks by fine-tuning the pre-trained model. 


\subsection{Representation Visualization and Retrieval}
\begin{figure}[t]
    \centering
    \begin{minipage}[4cm]{0.55\textwidth}
        \vspace{0.6cm}
        \centering
        \includegraphics[width=\textwidth]{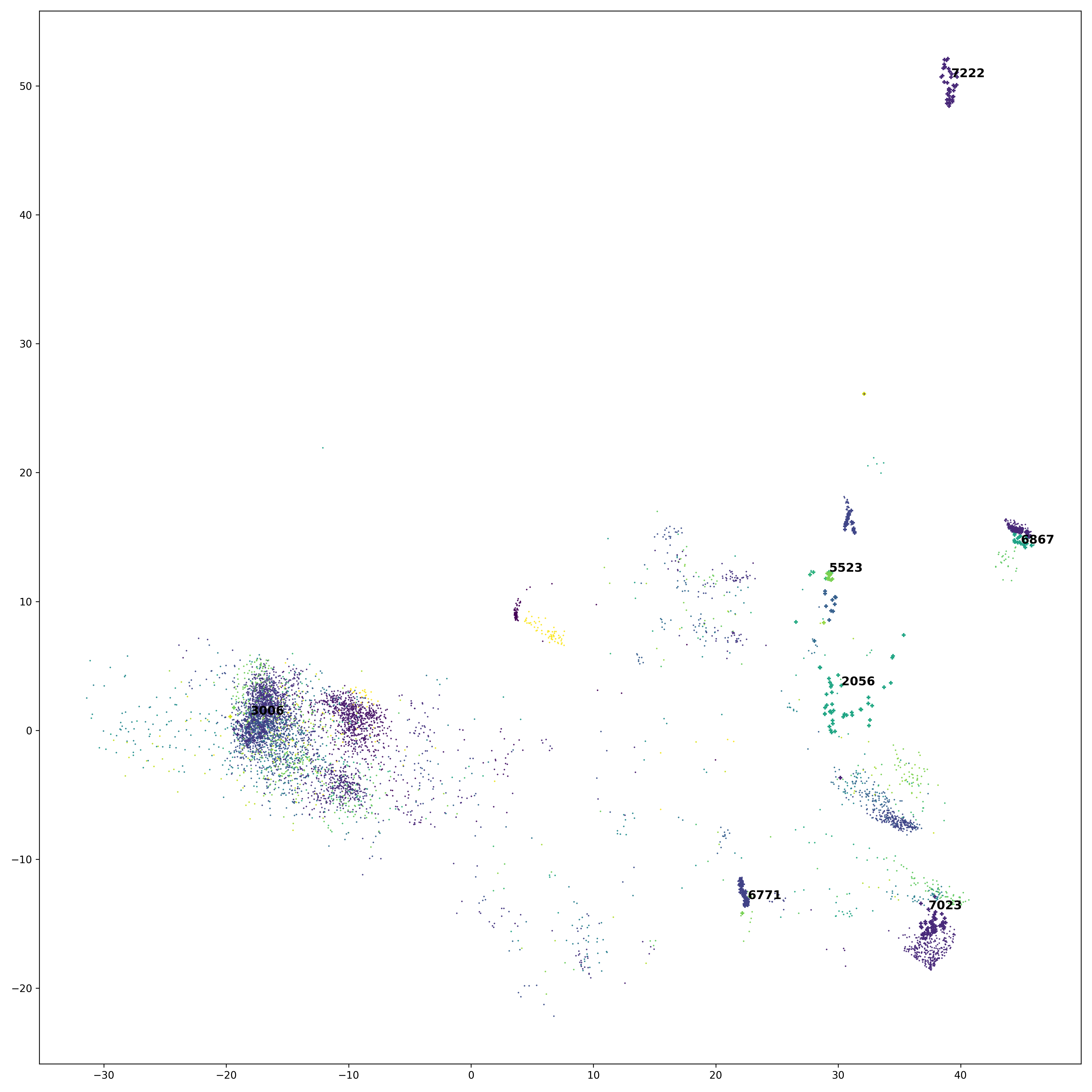}
        \caption{Visualization of the representation space learned by the MPPN on all MobIS cases. Different colors indicate different control-flow variants.}
        \label{fig:TSNE}
    \end{minipage}
    \hfill
    \begin{minipage}[4cm]{0.40\textwidth}
    \centering
    \captionsetup{type=table}
    \resizebox{\textwidth}{!}{
        \begin{tabular}{llrrrrr}
        \toprule
        \multirow{2}{1.1cm}{$c_{query}$} & \multirow{2}{1cm}{$\hat{\mathcal{C}}$ ID} & \multirow{2}{1.8cm}{$FV$ distance} & \multirow{2}{1cm}{DLD} & \multicolumn{3}{c}{MAE} \\
         & & & & \textit{timestamp} & \textit{cost} & \textit{travel\_start} \\
        \midrule
        \multirow{6}{*}{5523}
            &5511 &  0.00411 &           0 &        101.66 &      240 &                  0.53 \\
            &5613 &  0.00479 &           0 &        154.82 &      154 &                  6.47 \\
            &5036 &  0.01665 &           5 &       1307.83 &      244 &                 28.53 \\
            &5911 &  0.01755 &           8 &       1004.02 &      203 &                 24.47 \\
            &6034 &  0.01937 &           8 &        917.40  &      253 &                 31.93 \\
            &5980 &  0.02088 &           8 &        933.96 &      237 &                 28.55 \\
            &5868  &  0.02115 &            8 &       1045.91 &       69 &                 21.55 \\
        \midrule
        \multirow{6}{*}{2056}
            &4819 &  0.01388 &           0 &       2066.35 &       49 &                  174.00 \\
            &4960 &  0.02068 &           5 &       3587.52 &      218 &                181.33 \\
            &4765 &  0.02295 &           5 &        729.69 &      253 &                169.15 \\
            &4497 &  0.02340  &           5 &        632.51 &      217 &                  153.00 \\
            &4715 &  0.02428 &           5 &        717.51 &      263 &                  167.00 \\
            &5044 &  0.02453 &           5 &        847.96 &      375 &                  188.00 \\
            &4657  &  0.02465 &           5 &        689.45 &      233 &                162.39 \\
        \midrule
        \multirow{6}{*}{7222}
            &7109 &  0.00006 &           0 &         14.71 &       97 &                  7.35 \\
            &7092 &  0.00006 &           0 &         16.86 &       94 &                  8.42 \\
            &7073 &  0.00012 &           0 &         18.01 &       77 &                    9.00 \\
            &7090 &  0.00015 &           0 &         17.10  &       24 &                  8.54 \\
            &7133 &  0.00016 &           0 &         11.72 &       32 &                  5.86 \\
            &7231 &  0.00017 &           0 &          0.54 &      100 &                  0.27 \\
            &7052  &  0.00021 &           0 &         18.01 &       41 &                    9.00 \\
        \midrule
        \multirow{6}{*}{3006}
            &3227 &  0.00048 &           0 &         55.97 &      392 &                  153.00 \\
            &2403 &  0.00105 &           0 &        164.74 &     1123 &                    0.00 \\
            &2624 &  0.00118 &           0 &         54.72 &      501 &                   30.00 \\
            &3748 &  0.0012  &           0 &        206.65 &      662 &                  153.00 \\
            &2859 &  0.00123 &           0 &        103.17 &      629 &                   38.00 \\
            &2861 &  0.0014  &           0 &         89.77 &      474 &                   52.00 \\
            &2116  &  0.00153 &           0 &        287.39 &      250 &                   40.00 \\
        \bottomrule

        \end{tabular}}
        \caption{Similarities in the perspectives of the retrieved cases\\}
        \label{tab:retrieval}
    \end{minipage}
\end{figure}

In the following, we demonstrate how MPPN's internal representations $FV$ can be used for case-based case retrieval. Figure \ref{fig:TSNE} visualizes $FVs$ of each cases after they were reduced to a two-dimensional representation space using PCA. The training of the MPPN was performed analog to section \ref{approach:representation_learning} using the same input attributes as described in table \ref{fig:event_log}\footnote{We added $travel\_start$ as another attribute} but complete cases $c$ instead of prefixes. Note that the feature vectors hold information of all perspectives. Therefore, the clusters do not solely depend on the control-flow.

Figure \ref{fig:TSNE} shows that some clusters consist of cases with the same process variant. Other clusters are formed based on specific attribute combinations. For example, the biggest bulk shows all finished cases, i.e., complete cases from start to end containing the most common variant, represented by case 3006. One can make use of this representation for case-based case retrieval.  Given $L$ and a query case $c_{query}$, the task is to generate an ordered set of cases $\hat{\mathcal{C}}$ such that all cases in $\hat{\mathcal{C}}$ have similar characteristics as $c_{query}$. Instead of applying different filters on an event log to retrieve cases with particular characteristics, one can also retrieve cases starting with a specific case of interest. For this, the same feature vectors $FVs$ can now be used for retrieving such cases that share similar characteristics as a query case. First, the feature vector of the query case $FV_{query}$ is computed and compared to all other $FV$ of cases in $L$ using the cosine similarity. Next, the cases are sorted by their similarity, and those with the highest similarity are returned. 
We picked four cases as shown in table \ref{tab:retrieval} for retrieval and  marked the retrieved cases with bold symbols in figure \ref{fig:TSNE}. We see that the control-flow still is the deciding feature for the model as most of the retrieved cases have the same sequence of activities. Additionally, the retrieved cases have similar other characteristics as the query case:
\begin{itemize}
    \item 5523: Different process variants starting and ending with the same activities performed around the same date with cost below 1000.
    \item 2056: Cases that looped through the same activities with various number of this loop.
    \item 7222: Cases consisting of the first two events in the process performed around the same date with costs around 200.
    \item 3006: Complete cases from start to end of the most common variant.
\end{itemize}
From table \ref{tab:retrieval} one can see that the retrieved cases $\hat{\mathcal{C}}$ are similar in all perspectives to the query cases. We calculate the cosine distance of the $FV$, the Damerau-Levenshtein distance (DLD) and the mean absolute error for the three perspectives $\Pi_{cost}$,  $\Pi_{travel\_start}$ and $\Pi_{timestamp}$ (the MAE is computed after transforming the timestamps to durations).

\subsection{Next Step and Outcome Prediction}
This experiment evaluates the performance of the fine-tuned MPPN model in comparison to four baselines on the tasks next activity, last activity, next resource, last resource, event duration, and remaining time prediction.
\subsubsection{Datasets}
In this experiment, we consider seven event logs from different application domains.
The Helpdesk\footnote{http://doi.org/10.17632/39bp3vv62t.1} event log contains events from a ticketing management process of the help desk of an Italian software company.
Five event logs from the BPI Challenge 2012\footnote{https://doi.org/10.4121/uuid:3926db30-f712-4394-aebc-75976070e91f}. The original event log is taken from a Dutch Financial Institute and represents the application process for a personal loan or overdraft within a global financing organization. We included the original log as well as each sub-process individually.
The event log within BPI Challenge 2013\footnote{https://doi.org/10.4121/uuid:a7ce5c55-03a7-4583-b855-98b86e1a2b07} is an export from Volvo IT Belgium and contains events from an incident and problem management system called VINST.
The event log within BPI Challenge 2017\footnote{https://doi.org/10.4121/uuid:5f3067df-f10b-45da-b98b-86ae4c7a310b} is an updated, richer version of BPI Challenge 2012.
The event log from BPI Challenge 2020\footnote{https://doi.org/10.4121/uuid:52fb97d4-4588-43c9-9d04-3604d4613b51} was collected data from the reimbursement process at TU/e. We only included the request-for-payment log.
The MobIS event log\footnote{http://dx.doi.org/10.13140/RG.2.2.11870.28487} was elaborated in the MobIS Challenge \cite{Houy_2019}. It describes the execution of a business travel management process in a medium-sized consulting company.
We chose Helpdesk, BPIC 2012, and BPIC 2013 to achieve high comparability with existing approaches. BPIC 2017 and BPIC 2020 are selected as significantly more complex event logs that pose new challenges to prediction approaches while also revealing weaknesses of current approaches. MobIS contains several attributes and relationships, making it well-suited to demonstrate MPPN's multi-perspective approach's benefits. Table \ref{tab:event_logs} lists characteristics of each log and presents the attributes used as inputs for the process prediction tasks.

\begin{table}[t]
\centering
\caption{Event logs, statistics and attributes used}
\medskip
\label{fig:event_log}
\normalsize
\resizebox{\textwidth}{!}{
\begin{tabular}{lllC{2cm}C{2cm}L{5cm}L{3.7cm}L{1.7cm}}
\toprule
{} & 
\multirow{2}{*}{\#Traces} &  \multirow{2}{*}{\#Events} &  \multirow{2}{2cm}{\centering Avg. trace length} &  \multirow{2.7}{2cm}{\centering Avg. trace duration} & \multicolumn{2}{c}{Input Attributes }\\ \cmidrule(lr){6-8}&&&&&categorical&numerical&temporal
     \\
\midrule
Helpdesk   &     4580 &    21348 &               4.66 &           62.9 days &                                                                                            activity, resource &                                                                 &  timestamp \\
BPIC12     &    13087 &   262200 &              20.04 &          150.2 days &                                                                                            activity, resource &                                                      AMOUNT\_REQ &  timestamp \\
BPIC12\_Wc  &     9658 &    72413 &               7.50 &           95.6 days &                                                                                            activity, resource &                                                      AMOUNT\_REQ &  timestamp \\
BPIC13\_CP  &     1487 &     6660 &               4.48 &          426.5 days &  activity, resource, resource country, organization country, organization involved, impact, product, org:role &                                                                 &  timestamp \\
BPIC17\_O   &    42995 &   193849 &               4.51 &           23.9 days &                                                                     activity, Action, NumberOfTerms, resource &  FirstWithdrawalAmount, MonthlyCost, OfferedAmount, CreditScore &  timestamp \\
BPIC20\_RFP &     6886 &    36796 &               5.34 &           31.6 days &                                             org:role, activity, resource, Project, Task, OrganizationalEntity &                                                 RequestedAmount &  timestamp \\
MobIS      &     6555 &   166512 &              25.40 &         1194.4 days &                                                                                      activity, resource, type &                                                            cost &  timestamp \\
\bottomrule
\end{tabular}}
\label{tab:event_logs}
\end{table}

\subsubsection{Experimental Setup}
We compare MPPN with four different approaches \cite{Evermann_2017,Tax_2017,Camargo_2019,Pasquadibisceglie_2021}. For each task, the models receive as input case prefixes of increasing length, starting with the prefix that contains only the first event of a case up to the prefix that omits just the last event; i.e., for each case $\langle e_1, ..., e_n \rangle$  we create $n$ prefixes $\langle e_1, \dots, e_t \rangle$ with $0<=t<n$. In addition, we front-padded all prefixes to equal length.  
To make the results reproducible, we apply a random split between training, validation, and test cases for each data set, utilizing $20\%$ of the cases as test and $10\%$ of the cases as validation data. While the test set is fixed through all experiments and runs, the split between train and validation is performed randomly from run to run. All models were trained and validated with the same sets in each run. Each model was trained in the same fashion with a batch size of 512 while utilizing cyclical learning rates and early stopping \cite{Smith_2018}. 
The learning rate was picked with the learning rate finder algorithm as defined in \cite{Smith_2018}. Other than that, we picked the hyper-parameters of the baselines as mentioned in the corresponding papers. 
While \cite{Evermann_2017,Tax_2017,Camargo_2019} only considered control flow, resource and timestamp perspectives, the MiDA and the MPPN model is fed with all attributes listed in table \ref{fig:event_log}. We only removed attributes that contained duplicated information.
Last, we decided to remove all cases that are longer than 64 events since these are mostly outliers that falsify the prediction results and significantly increase training time. Each model was trained and tested ten times on all datasets and tasks.

\subsubsection{Prediction Tasks and Evaluation Metrics}
For this experiment, we formalize the prediction tasks and evaluation metrics as follows:\\
Given a prefix  $p_t = \langle e_1,..., e_t \rangle$  of a case $c= \langle e_1,..., e_n \rangle$ with $0<=t<n; t,n \in \mathbb{N}$, 
we define \textit{next step prediction} of an attribute $a$ as the task $NSP_a(p_t)$ that predicts $a(e_{t+1})$ based of the prefix $p_t$. 
We define \textit{outcome prediction} analogously to next step prediction as the task $OUT_a(p_t)$ that predicts $a(e_n)$ based on a prefix $p_t$. 
We measure the prediction performance of a model through a metric function, which is pairwise applied to all predictions and ground truth values for all prefixes over all cases and afterward combined to a final score. 
According to the type of the predicted attribute, it is necessary, to use different metric functions. In this experiment, we predict activity, resource, and timestamp. For activity and resource, we select the metric function accuracy. For timestamp, we convert it in the duration in days and then compute the mean absolute error.

\subsubsection{Baselines}
We re-implemented eight models as baselines from \cite{Evermann_2017,Tax_2017,Camargo_2019,Pasquadibisceglie_2021} based on the original papers and the corresponding source code. Our main objective is to reproduce the different network architectures, to be able to compare them in a fair and unified test setting with our MPPN.
\begin{table}[H]
\normalsize
\caption{Process prediction results}
\medskip
\resizebox{\textwidth}{!}{
\begin{tabular}{L{2.1cm} L{2.4cm} R{2.8cm} R{2.8cm} R{2.8cm} R{2.8cm} R{2.8cm} R{2.8cm}}
\toprule
 Dataset     & Model     &          $NSP_{activity}$
      &          $NSP_{resource}$ &          $OUT_{activity}$ &          $OUT_{resource}$ &          $NSP_{timestamp}$ &            $OUT_{timestamp}$ \\
\midrule
Helpdesk & Evermann\cite{Evermann_2017} &           0.651+-0.128 &           0.222+-0.005 &    \textbf{0.994+-0.000} &             0.811+-0.000 &           ---              &          ---              \\
      & Ca\_Spez.\cite{Camargo_2019} &           0.693+-0.168 &           0.289+-0.071 &    \textbf{0.994+-0.000} &             0.811+-0.000 &             7.95+-0.576 &             6.654+-0.101 \\
      & Ca\_concat\cite{Camargo_2019} &           0.696+-0.116 &           0.421+-0.035 &    \textbf{0.994+-0.000} &             0.811+-0.000 &             7.63+-0.052 &             6.739+-0.253 \\
      & Ca\_full\cite{Camargo_2019} &           0.774+-0.077 &             0.432+-0.000 &    \textbf{0.994+-0.000} &             0.811+-0.000 &            5.308+-0.288 &             7.018+-0.225 \\
      & Tax\_Spez.\cite{Tax_2017} &           0.763+-0.082 & ---                       &    \textbf{0.994+-0.000} &    ---                    &            7.777+-0.526 &             6.895+-0.253 \\
      & Tax\_Mixed\cite{Tax_2017} &             0.3+-0.003 &  ---                      &    \textbf{0.994+-0.000} &   ---                     &           14.849+-0.034 &             7.197+-0.101 \\
      & Tax\_Shared\cite{Tax_2017} &           0.793+-0.004 &  ---                      &    \textbf{0.994+-0.000} &     ---                   &            5.088+-0.129 &                6.67+-0.100 \\
      & MiDA\cite{Pasquadibisceglie_2021}&            0.693+-0.120 &           0.263+-0.089 &    \textbf{0.994+-0.000} &             0.811+-0.000 &   \textbf{4.898+-0.043} &    \textbf{6.629+-0.166} \\
      & MPPN &  \textbf{0.805+-0.003} &  \textbf{0.691+-0.006} &    \textbf{0.994+-0.000} &  \textbf{0.847+-0.008} &            5.197+-0.126 &             6.691+-0.089 \\ \midrule
BPIC12 & Evermann\cite{Evermann_2017} &           0.595+-0.107 &             0.149+-0.000 &             0.417+-0.000 &             0.172+-0.000 &         ---                &     ---                     \\
      & Ca\_Spez.\cite{Camargo_2019} &            0.795+-0.030 &           0.333+-0.282 &             0.417+-0.000 &           0.177+-0.015 &            0.693+-0.208 &              7.82+-0.033 \\
      & Ca\_concat\cite{Camargo_2019} &            0.74+-0.071 &           0.426+-0.164 &             0.417+-0.000 &             0.172+-0.000 &            0.722+-0.206 &             7.849+-0.076 \\
      & Ca\_full\cite{Camargo_2019} &           0.756+-0.064 &           0.283+-0.197 &             0.417+-0.000 &           0.184+-0.025 &            0.687+-0.226 &             6.649+-0.084 \\
      & Tax\_Spez.\cite{Tax_2017} &           0.585+-0.194 &  ---                      &             0.417+-0.000 &    ---                    &            0.734+-0.155 &             7.477+-0.127 \\
      & Tax\_Mixed\cite{Tax_2017} &           0.615+-0.182 &   ---                     &             0.417+-0.000 &    ---                    &            0.544+-0.205 &             6.678+-0.101 \\
      & Tax\_Shared\cite{Tax_2017} &           0.824+-0.008 & ---                       &           0.487+-0.019 &  ---                      &   \textbf{0.542+-0.167} &              6.693+-0.080 \\
      & MiDA\cite{Pasquadibisceglie_2021} &           0.565+-0.123 &             0.149+-0.000 &             0.417+-0.000 &             0.172+-0.000 &            0.625+-0.041 &    \textbf{6.587+-0.047} \\
      & MPPN &  \textbf{0.846+-0.006} &  \textbf{0.775+-0.002} &   \textbf{0.53+-0.005} &  \textbf{0.316+-0.004} &             0.82+-0.079 &             6.694+-0.066 \\ \midrule
BPIC12\_Wc & Evermann\cite{Evermann_2017} &             0.774+-0.000 &             0.104+-0.000 &             0.435+-0.000 &              0.11+-0.000 &            ---             &       ---                   \\
      & Ca\_Spez.\cite{Camargo_2019} &           0.775+-0.002 &             0.104+-0.000 &             0.435+-0.000 &           0.113+-0.005 &            1.799+-0.088 &              8.31+-0.058 \\
      & Ca\_concat\cite{Camargo_2019} &           0.794+-0.027 &             0.104+-0.000 &             0.435+-0.000 &           0.115+-0.013 &            1.843+-0.169 &             8.333+-0.041 \\
      & Ca\_full\cite{Camargo_2019} &           0.792+-0.026 &             0.104+-0.000 &           0.443+-0.026 &           0.112+-0.005 &             1.81+-0.125 &             7.455+-0.063 \\
      & Tax\_Spez.\cite{Tax_2017} &           0.713+-0.081 &   ---                     &             0.435+-0.000 &    ---                    &            1.765+-0.098 &             7.932+-0.086 \\
      & Tax\_Mixed\cite{Tax_2017} &             0.774+-0.000 &   ---                     &             0.435+-0.000 &    ---                    &   \textbf{1.595+-0.064} &              7.51+-0.106 \\
      & Tax\_Shared\cite{Tax_2017} &           0.773+-0.001 &   ---                     &           0.537+-0.057 &   ---                     &             1.645+-0.070 &     \textbf{7.409+-0.110} \\
      & MiDA\cite{Pasquadibisceglie_2021} &           0.805+-0.022 &             0.104+-0.000 &             0.435+-0.000 &  \textbf{0.155+-0.028} &            1.767+-0.115 &             7.424+-0.103 \\
      & MPPN &  \textbf{0.815+-0.006} &  \textbf{0.237+-0.011} &   \textbf{0.558+-0.01} &           0.147+-0.009 &            1.761+-0.061 &             7.528+-0.072 \\ \midrule
BPIC13\_CP & Evermann\cite{Evermann_2017} &           0.417+-0.113 &           0.082+-0.005 &      \textbf{1.0+-0.000} &             0.211+-0.000 &     ---                    &     ---                     \\
      & Ca\_Spez.\cite{Camargo_2019} &            0.481+-0.090 &             0.086+-0.000 &      \textbf{1.0+-0.000} &             0.211+-0.000 &           50.927+-2.339 &           137.718+-3.125 \\
      & Ca\_concat\cite{Camargo_2019} &           0.524+-0.004 &             0.086+-0.000 &      \textbf{1.0+-0.000} &             0.211+-0.000 &            51.672+-3.62 &           139.412+-5.032 \\
      & Ca\_full\cite{Camargo_2019} &           0.493+-0.065 &           0.106+-0.035 &      \textbf{1.0+-0.000} &             0.211+-0.000 &           67.168+-7.233 &            137.193+-6.280 \\
      & Tax\_Spez.\cite{Tax_2017} &           0.502+-0.067 & ---                       &      \textbf{1.0+-0.000} &    ---                    &           50.785+-4.395 &           140.481+-4.913 \\
      & Tax\_Mixed\cite{Tax_2017} &           0.309+-0.003 &  ---                      &      \textbf{1.0+-0.000} &    ---                    &          112.867+-0.279 &            176.167+-0.930 \\
      & Tax\_Shared\cite{Tax_2017} &            0.51+-0.011 &  ---                      &      \textbf{1.0+-0.000} &  ---                      &  \textbf{47.741+-1.217} &          144.528+-21.964 \\
      & MiDA\cite{Pasquadibisceglie_2021} &            0.434+-0.110 &           0.083+-0.005 &      \textbf{1.0+-0.000} &             0.211+-0.000 &           54.949+-4.044 &          128.185+-10.555 \\
      & MPPN &  \textbf{0.562+-0.009} &  \textbf{0.178+-0.024} &      \textbf{1.0+-0.000} &  \textbf{0.216+-0.008} &           54.922+-3.948 &  \textbf{127.824+-3.806} \\ \midrule
BPIC17\_O & Evermann\cite{Evermann_2017} &             0.818+-0.000 &           0.067+-0.005 &           0.509+-0.032 &           0.186+-0.041 &         ---                &     ---                     \\
      & Ca\_Spez.\cite{Camargo_2019} &             0.818+-0.000 &             0.064+-0.000 &           0.513+-0.018 &             0.192+-0.000 &            3.628+-0.057 &             9.604+-0.017 \\
      & Ca\_concat\cite{Camargo_2019} &             0.818+-0.000 &           0.226+-0.261 &           0.501+-0.027 &             0.192+-0.000 &            3.611+-0.082 &             9.606+-0.014 \\
      & Ca\_full\cite{Camargo_2019} &             0.818+-0.000 &           0.081+-0.048 &            0.52+-0.001 &             0.192+-0.000 &            3.627+-0.105 &             9.519+-0.025 \\
      & Tax\_Spez.\cite{Tax_2017} &            0.67+-0.065 &  ---                      &           0.454+-0.019 &     ---                   &            3.529+-0.019 &             9.688+-0.145 \\
      & Tax\_Mixed\cite{Tax_2017} &           0.726+-0.178 &   ---                     &           0.458+-0.014 &    ---                    &            3.999+-0.503 &             9.768+-0.184 \\
      & Tax\_Shared\cite{Tax_2017} &             0.818+-0.000 & ---                       &             0.519+-0.000 &  ---                      &            3.531+-0.037 &              9.47+-0.021 \\
      & MiDA\cite{Pasquadibisceglie_2021} &   \textbf{0.836+-0.030} &             0.064+-0.000 &  \textbf{0.828+-0.002} &             0.192+-0.000 &   \textbf{3.297+-0.037} &    \textbf{8.946+-0.059} \\
      & MPPN &             0.818+-0.000 &  \textbf{0.553+-0.061} &           0.518+-0.001 &  \textbf{0.208+-0.001} &            3.567+-0.068 &             9.534+-0.016 \\ \midrule
BPIC20\_RFP & Evermann\cite{Evermann_2017} &           0.699+-0.099 &           0.817+-0.084 &    \textbf{0.957+-0.000} &    \textbf{0.958+-0.000} &         ---                &    ---                      \\
      & Ca\_Spez.\cite{Camargo_2019} &           0.756+-0.087 &            0.841+-0.020 &    \textbf{0.957+-0.000} &    \textbf{0.958+-0.000} &            2.556+-0.142 &             6.068+-0.185 \\
      & Ca\_concat\cite{Camargo_2019} &            0.704+-0.09 &    \textbf{0.997+-0.000} &    \textbf{0.957+-0.000} &    \textbf{0.958+-0.000} &            2.631+-0.199 &             6.062+-0.079 \\
      & Ca\_full\cite{Camargo_2019} &           0.804+-0.025 &  \textbf{0.997+-0.001} &    \textbf{0.957+-0.000} &    \textbf{0.958+-0.000} &            2.634+-0.252 &             5.931+-0.117 \\
      & Tax\_Spez.\cite{Tax_2017} &           0.791+-0.085 & ---                       &    \textbf{0.957+-0.000} &   ---                     &            2.269+-0.085 &             5.933+-0.087 \\
      & Tax\_Mixed\cite{Tax_2017} &           0.431+-0.252 &  ---                      &    \textbf{0.957+-0.000} &   ---                     &            3.827+-2.194 &              8.55+-0.058 \\
      & Tax\_Shared\cite{Tax_2017} &  \textbf{0.849+-0.001} &  ---                      &    \textbf{0.957+-0.000} &   ---                     &    \textbf{2.12+-0.095} &    \textbf{5.468+-0.181} \\
      & MiDA\cite{Pasquadibisceglie_2021} &            0.55+-0.109 &  \textbf{0.997+-0.001} &    \textbf{0.957+-0.000} &    \textbf{0.958+-0.000} &            2.673+-0.173 &             5.842+-0.086 \\
      & MPPN &  \textbf{0.849+-0.001} &    \textbf{0.997+-0.000} &    \textbf{0.957+-0.000} &    \textbf{0.958+-0.000} &            3.018+-0.849 &             6.495+-0.909 \\ \midrule
MobIS & Evermann\cite{Evermann_2017} &            0.767+-0.140 &             0.163+-0.000 &             0.798+-0.000 &             0.075+-0.000 &         ---                &      ---                    \\
      & Ca\_Spez.\cite{Camargo_2019} &             0.87+-0.040 &             0.163+-0.000 &             0.798+-0.000 &             0.075+-0.000 &             4.648+-0.560 &            30.106+-0.814 \\
      & Ca\_concat\cite{Camargo_2019} &           0.836+-0.034 &             0.163+-0.000 &             0.798+-0.000 &             0.075+-0.000 &            4.801+-0.525 &            30.133+-0.526 \\
      & Ca\_full\cite{Camargo_2019} &           0.838+-0.038 &             0.163+-0.000 &             0.798+-0.000 &             0.075+-0.000 &            3.966+-0.922 &            24.449+-0.354 \\
      & Tax\_Spez.\cite{Tax_2017} &            0.85+-0.079 &  ---                      &             0.798+-0.000 &    ---                    &            3.919+-0.968 &            28.236+-1.569 \\
      & Tax\_Mixed\cite{Tax_2017} &           0.545+-0.188 & ---                       &             0.798+-0.000 & ---                       &            2.333+-0.602 &            21.384+-0.977 \\
      & Tax\_Shared\cite{Tax_2017} &           0.926+-0.008 & ---                       &           0.805+-0.009 &  ---                      &   \textbf{2.323+-0.638} &    \textbf{20.963+-0.420} \\
      & MiDA\cite{Pasquadibisceglie_2021} &             0.7+-0.154 &             0.163+-0.000 &             0.798+-0.000 &           0.075+-0.003 &            2.992+-0.372 &            24.498+-0.405 \\
      & MPPN &  \textbf{0.934+-0.003} &  \textbf{0.536+-0.026} &  \textbf{0.812+-0.002} &  \textbf{0.121+-0.023} &             4.827+-0.420 &            22.454+-1.011 \\
\bottomrule
\end{tabular}}
\label{tab:benchmark}
\end{table}

\noindent Thus, we deviate from the original work in some aspects regarding train-test splitting, sequence generation, and pre-processing, which also leads to different prediction results. Unfortunately, we cannot guarantee that we have correctly reproduced all the details of the specifications of the models, due to missing source code, documentation or test data. 


\subsubsection{Interpretation of Results}
For the final comparison, we averaged the prediction scores over ten runs.  Table \ref{tab:benchmark} presents the final results. There is no superior model that performs best in all tasks on all datasets. However, the results suggest the effectiveness of the MPPN. 
This yields in particular for the
$NSP_{activity}$ and the $NSP_{resource}$ tasks, where it achieves the highest scores on nearly all datasets.
The MPPN also performs well on the $OUT_{activity}$ and the $OUT_{resource}$ tasks. However, there is not such a wide performance variety between the models. Most of the examined processes only have a few outcome classes. Therefore, the tasks are supposed to be simpler and lead to similar results. At the same time, the available information in the prefixes may not always allow for a adequate prediction. 
For the two regression tasks, the MPPN achieves solid but no outstanding results.
Overall, the results suggest that the MPPN model is more robust than the other models and does not require extensive hyperparameter tuning.
One explanation might be that the MPPN utilizes gramian angular fields in combination with CNNs instead of embeddings and recurrent layers. Also, the CNN in MPPN is based on the Alexnet architecture, which has been carefully optimized for image recognition tasks. 
\cite{Tax_2017,Camargo_2019,Pasquadibisceglie_2021} utilize multi-task learning without fine-tuning, which seem to fail occasionally to optimize one particular task fully. In contrast, through the fine-tuning step of the MPPN, it can focus on one task at a time. Additionally, the MPPN performs reasonable overall tasks and datasets which is a strong indicator that it can learn effective, general representations of the underlying process.
Another interesting aspect is the influence of the different perspectives on the process predictions. The MPPN and the MiDA model utilized almost all available perspectives, while the other models only examined activity, resource, and timestamp. In the datasets containing contextual attributes, the MPPN can often outperform other methods indicating that the model can make use of the additional information and embed them into the representation.  In the future, we plan to further investigate the influence of different datasets and subsets of perspectives. For example, in the case of BPI17, we expect that contextual information such as application type and event origin can positively affect the prediction quality.

\subsubsection{Reproducibility} \label{sec:reproducability}
All code used for this paper, including the implementation of MPPN as well as the case retrieval and the prediction experiments, can be found in our git repository\footnote{http://bit.do/fQRbF}.

\section{Conclusion}
In this work, we have proposed a novel approach for multivariate business process representation learning utilizing gramian angular fields and convolutional neural networks. MPPN is a generic method that generates multi-purpose vector representations by exposing important characteristics of process instances without the need for manually labeled data. We showed how these representations can be exploited for analytics tasks such as clustering and case retrieval. Furthermore, our work demonstrated the advantages of meaningful, general representations for later downstream tasks such as next step and outcome prediction. In the performed experiments, we were able to outperform existing approaches and generate robust results over several datasets and tasks. This demonstrates that representation learning can successfully be applied on business process data. Furthermore, the self-supervised pre-training makes the model robust and helps in cases where contextual information is given. Additionally, in spite of recent advances in NLP, our result indicate that a non-recurrent neural network outperforms other architectures that use recurrent layers. 

One limitation of this paper is a missing systematic hyper-parameter tuning. In this paper, we investigated the robustness of the models on multiple datasets and tasks making it a generic approach. In the future, we want to elaborate on how hyper-parameter tuning can improve the performance of a specific model on a given dataset. 
Furthermore, we plan to investigate how the approach can explain the impact of certain attributes on other events in a process. The "black box" nature of deep learning models is still a major issue in the context of predictive process analysis.
Last, we want to elaborate more approaches and ideas from other domains such as natural language processing and computer vision to learn richer representations capturing more and finer characteristics.

\bibliographystyle{splncs04}
\bibliography{bib}
\end{document}